\documentclass[10pt,twocolumn,letterpaper]{article}

\usepackage[pagenumbers]{cvpr} 
\usepackage{longtable}
\usepackage{float}
\usepackage{subcaption}
\usepackage{booktabs}
\usepackage{multicol}
\usepackage{tikz}
\usepackage{media9}
\usepackage{tabularx}
\usepackage{enumitem}
\newtheorem{definition}{Definition}
\usepackage[load-configurations=version-1]{siunitx} 

\definecolor{cvprblue}{rgb}{0.21,0.49,0.74}
\usepackage[pagebackref,breaklinks,colorlinks,allcolors=cvprblue]{hyperref}

\def\paperID{14392} 
\def\confName{CVPR}
\def\confYear{2025}

\title{Learning Artistic Signatures: Symmetry Discovery and Artistic Style}

\author{Emma Finn \\
The Kemnpner Institute for the Study of Natural and Artificial Intelligence at Harvard University\\
{\tt\small efinn@college.harvard.edu}
\and
T. Anderson Keller \\
{\tt\small t.anderson.keller@gmail.com}
\and 
Manos Theodosis \\
{\tt\small etheodosis@g.harvard.edu}
\and
Demba E. Ba \\
{\tt\small demba@seas.harvard.edu}
}

\begin{document}
\maketitle
\begin{abstract}
Despite nearly a decade of literature on style transfer, there is no undisputed definition of artistic style. State-of-the-art models produce impressive results but are difficult to interpret since, without a coherent definition of style, the problem of style transfer is inherently ill-posed. Early work framed style-transfer as an optimization problem but treated style as a measure only of texture. This led to artifacts in the outputs of early models where content features from the style image sometimes bled into the output image. Conversely, more recent work with diffusion models offers compelling empirical results but provides little theoretical grounding. To address these issues, we propose an alternative definition of artistic style. We suggest that style should be thought of as a set of global symmetries that dictate the arrangement of local textures. We validate this perspective empirically by learning the symmetries of a large dataset of paintings and showing that symmetries are predictive of the artistic movement to which each painting belongs. Finally, we show that by considering both local and global features, using both Lie generators and traditional measures of texture, we can quantitatively capture the stylistic similarity between artists better than with either set of features alone. This approach not only aligns well with art historians' consensus but also offers a robust framework for distinguishing nuanced stylistic differences, allowing for a more interpretable, theoretically grounded approach to style transfer.
\end{abstract}    

\section{Introduction}
\label{sec:intro}
\label{sec:intro} What makes a Van Gogh a Van Gogh? Maybe the blue-green color palette or his distinctive swirling skies. Whatever it is, it does not translate neatly into numeric terms. This problem has complicated the field of style transfer since its inception in the early 2010s. When Gatys et al. first published A Neural Algorithm of Artistic Style in 2015, they hoped to offer “a path forward to an algorithmic understanding of how humans create and perceive artistic imagery.”~\cite{gatys2016} Subsequent work has mostly fallen short of this goal, in part because no one can agree on what constitutes style. Two branches of work have developed from Gatys’ seminal paper, namely (I) the “optimization school,” which broadly works to optimize a loss function defined in terms of features extracted from the content image and textures extracted from the style image; and (II) the “probabilistic school”, primarily using pre-trained diffusion models with text embeddings to map the latent representation of the content image to the embedding of the style image. \cite{Zhang_2023_CVPR, domain_enhanced, style_injection}

\begin{figure*}[h!]
    \centering
    \includegraphics[width=.90\linewidth]{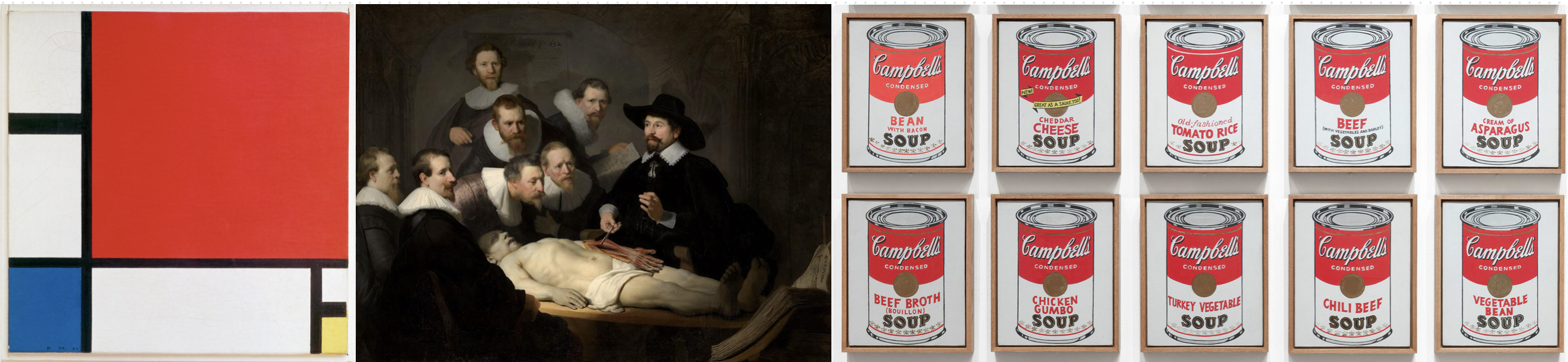}
    \caption{Mondrian's Composition ii (left), Rembrandt's The Anatomy Lesson (middle), Warhol's Campbell's Soup Cans (right)}
    \label{fig:mondrian}
\end{figure*}
We propose and defend a mathematically grounded notion of style that combines the local, textural approach of the optimization school with a group-theoretic approach to the arrangement of global features. This accords with human intuition for what separates, for instance, a Rembrandt from a Mondrian (Figure \ref{fig:mondrian} middle \& left respectively). Rembrandt’s chiaroscuro characterizes his work just as much as the geometric arrangement of the figures and their relative orientations. Mondrian’s composition, on the other hand, is defined by the unnaturally bright colors and sharp angles. Similarly, Andy Warhol's soup cans are distinguished from both by the hyper-rigid global structure of repeating images, which would not be captured by simple local textural analysis. We suggest that style can be thought of as a set of local features (including texture) with a global arrangement described by a symmetry group. 

We provide evidence for the utility of this definition through experiments that highlight the importance of including symmetry in any notion of style. Our methods build on literature by Daubechies and Shamir, which used older, more standard statistical methods to group paintings by similarity. \cite{shamir_2012, Daubechies2012}  In particular, \citet{shamir_2012}, using a smaller dataset not dissimilar to ours, quantifies the similarity of paintings across more than 4,000 pre-determined, hand-picked features and demonstrates that “computer analysis can automatically associate different painters based on their artistic styles.” We modernize their approach, applying a recently developed symmetry discovery technique, LieGG  \cite{moskalev2022liegg}, and demonstrate through analogous experiments that the inferred symmetry group generators have the same capacity as the prior hand-crafted image features to distinguish artists and movements across history. 

In further support of our definition, we emphasize that in the disentanglement literature, symmetries have long been used to separate style from content. A major goal of the field has been to learn representations that precisely factorize according to the underlying `generative factors', or transformations, of the data. In ``Towards a  Definition of Disentangled Representations'', \citet{higgins2018definitiondisentangledrepresentations} suggest that representing objects via their ``transformation properties'' is the most natural and useful representation for generalization and downstream performance. In the equivariant neural network literature, \citet{cohen2014} have similarly argued that the ``irreducible representations'' of the symmetry groups of a dataset are the most natural ``elementary components'' of said data. Though the implementation details differ, the spirit is the same: 
symmetries capture the invariant transformation properties of objects, and therefore provide a principled way to separate out sources of dataset variation. 
Borrowing  from this work, in the following, we propose a definition of artistic style based on the symmetries of a painting and demonstrate that this definition aligns both with human intuition and art-historical expertise.

\section{Related Work}
Before introducing our new method, we overview prior work on the definition of artistic style, primarily developed in the domain of style transfer, initially by members of the so-called Optimization School and more recently by the Probabilistic School. 
\label{sec:related_work}
\subsection{The Optimization School}
At a high level, work in the optimization school seeks to minimize the weighted sum of content and style losses, which are computed by passing the content image, style image, and a noisy target image through a trained classifier network, acting as a feature extractor. The resulting feature maps from the target image are then compared to the feature maps from the style and content images at various layers. Explicitly, this combined loss is given as
\begin{equation}
    \mathcal{L}_{\text{total}}(\vec{p}, \vec{a}, \vec{x}) = \alpha\mathcal{L}_{\text{content}}(\vec{p}, \vec{x}) + \beta\mathcal{L}_{\text{style}}(\vec{a}, \vec{x}),
\end{equation}
where $\vec{x}$ is the target image, $\vec{p}$ is the content image, $\vec{a}$ is the style image, and 
$\alpha, \beta \in \mathbb{R}$ are weights that allow the user to adjust the relative fidelity to the style or content images.

The standard formulations of the content loss is 
\begin{equation}
    \mathcal{L}_{\text{content}}(\vec{p}, \vec{x}) = \frac{1}{2} \sum_{l =0}^L w_l^c \sum_{i,j} (F_{ij}^l(\vec{x}) - F_{ij}^l(\vec{p}))^2,
\end{equation}
where $F_{ij}^l(\vec{x})$ is the activation of the $i$th filter at position $j$ and layer $l$ of the target image, and $F_{ij}^l(\vec{p})$ is the analogous set of feature maps of the content image. Traditionally the set of content layers with non-zero weights $w_l^c$ are the last convolutional layers in the classifier, since those are thought to represent higher level, global arrangements of features. 

The style loss is computed by taking a weighted average of the difference in ``textures'' across the earlier convolutional layers of the network. These textures are encapsulated in a `Gram Matrix' $G^l$ which is the pair-wise inner-products of convolutional feature maps for all learned convolutional filters of a given layer. Explicitly, 
\begin{equation}
    \label{eqn:gram}
    G^l_{ij}(\vec{x}) = \sum_h \sum_w F^l_{ihw}(\vec{x})F^l_{jhw}(\vec{x})
\end{equation}
where $F^l(\vec{x}) \in \mathbb{R}^{N_l \times H_l \times W_l}$ is the set of $N_l$ feature maps for the $l$th convolutional layer, each with spatial height $H_l$ and width $W_l$. The full style loss is then given as 
\begin{equation}
    \label{eqn:style_loss}
    \mathcal{L}_{\text{style}}(\vec{a}, \vec{x}) = \sum_{l=0}^L w_l^s \frac{1}{4 N_l^2 M_l^2} \sum_{i,j} \left( G_{ij}^l(\vec{x}) - G_{ij}^l(\vec{a}) \right)^2,
\end{equation}
where the inner sum computes the difference between the Gram matrices of the content and style images, and the outer sum then weights these layer-wise contributions with $w^s_l$.

While subsequent work has made meaningful computational strides and improved the quality of the output, authors such as \citet{perceptual2016} acknowledge that the problem of style transfer remains ill-posed. In general, these follow-up papers offer major efficiency improvements and processing tweaks but still leverage the Gram matrix as the primary measure of style when its original intention was as a texture metric. \cite{gatys2015, gatys2016} 
While this optimization approach creates images that are visually satisfactory \cite{universal, wright2022artfid}, they don’t move us any closer to the original goal of understanding “how humans create and perceive artistic imagery.” \cite{gatys2016neural}  

\subsection{The Probabilistic School}
Recently, probabilistic approaches to style transfer have achieved extraordinary results but fail to justify those results with any comprehensive definition of style, which hinders interpretability. Diffusion models aim to learn the underlying probability distribution for a dataset by stochastically diffusing samples towards a known simpler distribution, such as a standard Gaussian, and then learning an approximate inverse process to sample from the data distribution. 
To modify this procedure for style transfer, the stochastic process is often conditioned on a text embedding of a particular style in order to learn the conditional image distribution. The methods here are more diverse than those applied in the optimization school. Importantly, however, these models lack formal inductive biases for style, relying instead on learned representations of text prompts, sometimes with reference images.

As a consequence of this `black-box' approach, such models tend to reflect texture less accurately; so while color, structure, and resolution are more faithful to the style image, they lack some of the qualities that made Gatys et al.'s work so visually striking. One particularly exciting recent method, "StyleDiffusion" \cite{stylediffusion}, attempts to disentangle style from content by first extracting the “content information and then explicitly learn[ing] the complementary style information.” This method crucially prevents over-reliance on textural components of style, but doesn’t provide any generalizable framework for style other than asserting that it is “everything that is not content.” Other articles, including Dreambooth~\cite{ruiz2022dreambooth} and LaDiffGan~\cite{Liu_2024_CVPR}, generate content “prototypical” of a particular artist or movement,rather than displaying style transfer which aligns with abstract art historical notions. In "Style Injection" \cite{style_injection}, the authors meld the ideas of prior work, arguing that their method employs cross-attention to capture the key features of the style image. In particular they substitute the key-value pairs corresponding to the cross-attention maps from the style image into the diffusion process for generating a stylized image. 

Despite these advances, as in prior work, the fundamental issue remains unaddressed. This probabilistic approach to restyling images relies primarily on good results rather than theoretical groundwork and, in so doing, avoids the most interesting and challenging component of style transfer: reliably separating style from content. 

\section{A Proposed Definition}
We suggest that artistic style is best defined in terms of a combination of symmetry transformations (the `fundamental elements' of global structure), and local textural features (the `feeling' of local elements). To define this more precisely, we review continuous symmetry groups, explain why this structure cannot be captured tractably by a Gram-type metric, and provide an alternative computable metric that can be used to capture our definition. In Section \ref{sec:methods}, we validate this proposed definition through comparison with known art-historical categories. 


\subsection{Lie Algebra Generators}
To begin, we first review the definitions of Lie groups, their associated algebras, and generators, from Lee's "Introduction to Smooth Manifolds" \cite{lee2013introduction}. \footnote{We refer interested readers to the full text for a complete introduction.} 

\begin{definition}
A Lie group is a smooth manifold G which also has the algebraic structure of a group. 
\end{definition}

\begin{definition}
A Lie algebra $\mathfrak{g}$ over $\mathbb{R}$ is a vector space endowed with a map, called the bracket from $\mathfrak{g} \times \mathfrak{g} \rightarrow \mathfrak{g}$, which is bilinear, anti-symmetric, and obeys the Jacobi identity.
\end{definition}

\begin{definition}
The generators $\mathfrak{h_1} ... \mathfrak{h}_k \in \mathfrak{g}$ of a Lie group G are the basis elements of the group's Lie algebra. 
\end{definition}
Intuitively the Lie group encapsulates the set of continuous symmetries to which a set of objects is invariant. For instance, Warhol's Soup cans in Figure ~\ref{fig:mondrian} are highly translation invariant and display lots of horizontal and vertical symmetries. The Lie algebra is the tangent space to that manifold at the identity and the generators correspond to ``infinitesimal'' group transformations near the identity. In our circumstance, our Lie bracket is given by the exponential map which sends elements of the Lie algebra to the Lie group. 
Thus, if we can identify the basis elements of our Lie algebra, we can describe all elements of the symmetry group $\mathcal{G}$, to which our dataset is invariant. We will use these generators to stand for the symmetries which characterize a given painter's work.  

\subsection{Gram Matrices}
As described above, Gram matrices describe the texture of an image. The Gram matrix consists of all pair-wise inner-products of convolutional feature maps for all learned convolutional filters of a given layer. As given in Equation \ref{eqn:gram}, $G^l_{ij}$ therefore captures the co-occurrence of features $i$ and $j$ at roughly the same spatial location, but invariant to where the location is precisely. 

Clearly, $G^l$ is useful for representing texture, since it sums over all spatial dimensions resulting in a translation invariant measure; however, as a result, it cannot capture information about larger scale global arrangement. For example, if we consider a what $G^l$ might look like for the Mondrian piece in Figure \ref{fig:mondrian} (left), it would likely capture the complete lack of texture, but it is highly unlikely to capture the fact that the entire image may be representable by scaling, translation, and color transformations, applied to a single rectangular shape (\textit{i.e.} the global symmetry structure). To be able to capture this more complex co-occurrence structure across different spatial locations, scales, and colors, one would need to preserve the corresponding transformation group indices in the matrix $G$, and instead compute a sum over the irrelevant feature indices. 

Explicitly, if we were to define the transformations of aspect-ratio scaling as $s \in \mathcal{G}_s$, color transformation as $c \in \mathcal{G}_c$, and 2D translations as $t \in \mathcal{G}_t$, we could theoretically capture the fact that a given feature is present at a variety of different group element `locations' simultaneously with:
\begin{equation}
    G^l_{s \hat{s} c \hat{c} t \hat{t}}(\vec{x}) = \sum_{i} F^l_{i s c t}(\vec{x})F^l_{i \hat{s} \hat{c} \hat{t}}(\vec{x}),
\end{equation}
where, $F^l_{i s c t}$ corresponds to an organized feature map such that each filter $i$ has a set of scales $s$, color transformations $c$, and locations $t$. Such a Gram matrix would have large positive values corresponding to the interactions between scales, colors, and relative positions, exactly as those in the Mondrian work, but agnostic to the underlying features themselves. Unfortunately however, as can be seen simply from the equation above, this formulation not only requires that the feature maps be neatly organized in terms of the generating group elements (requiring something like a Group Equivariant Neural Network \cite{cohen2016}), it also requires that the corresponding representation of style ($G^l_{s \hat{s} c \hat{c} t \hat{t}})$ grow exponentially in size with the group dimensions. 

Given that this idealized Gram matrix (or tensor) is significantly more computationally complex than the original formulation, and therefore may not be efficiently computable, we instead seek a more tractable combination of the original texture-based Gram matrix, and symmetries defined by the more compact Lie group generators. In the following, we describe how to accomplish this.

\subsection{A Combined Metric}
In order to create a definition of style which captures both local texture and the global arrangement of features, we propose to augment the Gram matrix representation of style with the Lie group generators discovered for a given artist or artistic movement. 

At a high level, the original style loss metric in Equation \ref{eqn:style_loss} can be thought of as measuring a distance between textures of two images, in this case the target and style images. In our work, we propose that this distance may be abstracted to encompass two arbitrary `texture' description matrices, $G$ and $G'$, potentially representing a given painting, artist, or artistic movement as a whole. This distance can then be written as $d_{\text{texture}}(G, G')$. In Section \ref{sec:gram_distance}, we outline precisely how we compute this distance and the associated matrices $G$ for defining an artist's style. 
Analogously, we then propose to introduce an additional distance metric, $d_{\text{global}}(\mathfrak{g}, \mathfrak{g}')$ between the associated Lie algebras for two paintings or bodies of work. In Section \ref{sec:methods}, we outline precisely how we can compute such generators, and further how we can compute an accurate distance between two such vector spaces. 
Given these two distances, we can then return to, and formalize, our original suggestion that style is the arrangement of a finite set of local, textural features arranged globally according to a symmetry group. 

\begin{definition}
    Given a set of images $\{\vec{x}_i\}_{i \in \mathcal{S}}$, which are said to have style $s$,  
    an image $\vec{x}_j$, can be said to also have style $s$ if (I) $\vec{x}_j$ has a Gram matrix $G(\vec{x}_j)$ that is within some tolerance $\epsilon_1$ of the average Gram matrix $\bar{G}_s$ computed from the images of style $s$, i.e. $d_{\text{texture}}(G({\vec{x}_j}), \bar{G}_s) < \epsilon_1$, and (II) $\vec{x}_j$ has Lie generators $\mathfrak{g}_{\vec{x}_j}$ which are within some tolerance $\epsilon_2$ of the generators $\mathfrak{g}_s$ of the set of images $\{\vec{x}_i\}_{i \in \mathcal{S}}$ of style $s$, i.e. $d_{\text{global}}(\mathfrak{g}_{\vec{x}_j}, \mathfrak{g}_s) < \epsilon_2$.
\end{definition}
Finally, to achieve a relaxation of this definition, we propose that a single unified metric of style may be composed of a convex combination of textural and symmetry distances as:
\begin{equation}
    \label{eqn:combined}
    (1 - \lambda) d_{ \text{texture}}(G(\mathbf{x}), \bar{G}_s) + \lambda d_{ \text{global}}(\mathfrak{g}_{\vec{x}}, \mathfrak{g}_s).
\end{equation}

While this combined metric is conditional on the specification of the set $\{\vec{x}_{i}\}_{i \in \mathcal{S}}$ which defines the style, in the remainder of this work, we will show how by defining these sets per-artist, a weighted combination of the above metrics yields a strongly structured representational space which matches both art-historical notions of artistic movements, as well as prior work on computational metrics of style. \cite{shamir_2012}


\section{Methods}
\label{sec:methods}
In this section, we outline our proposed method for tractably computing the features and distances required to evaluate the combined metric in Equation \ref{eqn:combined} and our approach to experimentally verifying it.

\subsection{Computing the Lie Generators}
In \citet{moskalev2022liegg}, the authors offer a practical method, named LieGG, for computing the infinitesimal Lie group generators that a given neural network has learned to be invariant with respect to. In this work, we apply this technique to a complex dataset for the first time. First, we train a suite of Multi-Layer Perceptrons (MLPs) for binary classification of images according to their respective artists. 
We train 50 of these 4 layer networks, with 384 units per layer, one for each artist. Then, we solve for the group elements $g \in \mathcal{G}$ which the network $f_{\theta}$ is invariant to. In particular, as detailed in the LieGG paper, if the network is invariant to the action of $g \in \mathcal{G}$, it follows that 
\begin{equation}
    f_{\theta}(g \curvearrowright \vec{x}) = 0 \iff f_{\theta}(\vec{x}) = 0
\end{equation}
where $ g \curvearrowright \vec{x}$  denotes the action of a group element $g$ on a painting $\vec{x}$. 
Then, we use the local linearity of the Lie algebra to solve for the generators of such $g \in \mathcal{G}$. In particular, for some generator $\mathfrak{h} \in \mathfrak{g}$ we Taylor expand to find,
\begin{equation}
f_\theta(\vec{x} + \epsilon \, \mathfrak{h} \cdot \vec{x}) \approx f_\theta(\vec{x}) + \epsilon \sum_{i=1}^n \frac{\partial f_\theta}{\partial x_i} (\mathfrak{h} \cdot \vec{x})_i
\end{equation}

Observe that since we require $f_{\theta}(\vec{x}) = 0$, the infinitesimal symmetry transformation must leave the output of the model approximately zero, so 
\begin{equation}
\sum_{i=1}^n \sum_{j=1}^n \frac{\partial f_{\theta}}{\partial x_i} \cdot \mathfrak{h}_{ij} \cdot x_j = 0.
\end{equation}
Solving this system of linear equations for $\mathfrak{h}_{ij}$ gives us our symmetry generators. We typically solve this by Singular Value Decomposition (SVD), which has the additional benefit that the generators are listed in order of decreasing singular values, representing in some sense the ordering of transformations to which the network is ``most equivariant.'' We select the top four estimated elements of the Lie algebra to be our symmetry style description of a given dataset.

\subsection{Computing Distances Between Lie Algebras}
\label{sec:lie_distance}

Once we have these elements, computing the ``difference'' in their symmetries amounts to computing the distance between the subspaces spanned by those top four elements. We're interested in how ``close'' the symmetries are, which corresponds to the distance between group elements on the smooth manifold, rather than the distance between the generators in the algebra. Since we take the top four Lie generators from our painters, we consider the subspace spanned by those four generators. For all painters, these have the same dimension, so we use the so called ``Grassman Distance'' to compute the deviation between them, by measuring the principal angles between them. For a $k$-dimensional subspace, these principal angles $\theta_1, ..., \theta_k \in [0, \frac{\pi}{2}]$ represent the primary directions of symmetry for each painter, where smaller angles suggest that the subspaces are more similar and thus that the artists share more artistic qualities \cite{Hamm2008Grass}. 

The principal angles are computed as follows. First, we define the principal vectors $a^*_j$ \& $b^*_j$, and angles $\theta_j$:

\begin{definition}
    Let $A, B$ be two subspaces of dimension $k$ in $\mathbb{R}^n$ spanned by $a_1, ..., a_k$ and $b_1, ..., b_k$ respectively. The principal vectors $(a_j^*, b_j^*) \in A \times B$  are defined recursively as the solutions to maximizing  $a^T b$
subject to:
\[
a \in \text{span}(a_1, \ldots, a_{j-1}), \quad b \in \text{span}(b_1, \ldots, b_{j-1}),
\]
\[
a^T a_i = b^T b_i = 0  \quad \forall i < j, \quad \|a\| = \|b\| = 1,
\]
for \( j = 1, \ldots, k \).
\end{definition}

\begin{definition}
    The principal angles $\theta_j$ are defined to be $\theta_j = \arccos({{a_j^*}^T}b_j^*$) for $j \in \{1,..,k\}$
\end{definition}

These subspaces $A,B \in \mathbb{R}^n$ are elements of the Grassman manifold $Gr(k,n)$ \cite{Hamm2008Grass}, which admits the following definition and distances:
\begin{definition}
    The Grassman manifold $Gr(k,n)$ is the set of all $k$-dimensional subspaces of $\mathbb{R}^n$.
    The Grassman distance is then given as $d_{Gr}(A,B) = \left( \sum_{i = 1}^{k} \theta_i^2\right)^{\frac{1}{2}}$.
\end{definition}

For each painter, we compute an orthonormal basis to ensure that each angle $\theta_j$ represents unique information about similarity along a direction of symmetry.~\cite{Ye2016Grass} This method respects the structure of the underlying Lie group, which ensures that our metric measures the distance between elements of the symmetry group. In practice, this is accomplished by 
computing the cross product $M = A^TB$, and performing singular value decomposition $M = U \Sigma V^T$, to extract the diagonal matrix $\Sigma$, containing as its diagonal entries $\sigma_i = \cos(\theta_i)$. We compute this distance pair-wise for each artist, and end with a $50\times50$ matrix representing the distances between each pair of artists. 

\subsection{Distance Between Gram Matrices}
\label{sec:gram_distance}
To compare the textural features of our images, we compare their Gram matrices. In particular, we compute the average Gram matrix for each painter, by passing our labeled images through an ImagNet pretrained VGG-19 convolutional neural network \cite{vgg}, extracting the feature maps at convolutional layers $\text{conv}1_1, \text{conv}2_1, \text{conv}3_1, \text{conv}4_1$, 
and then computing the average Gram matrix for artist $k$ as 
\[
\bar{G}_{k}= \frac{1}{|\mathcal{S}_k|} \sum_{i \in \mathcal{S}_k} G(\vec{x}_i)
\]
where \( \mathcal{S}_k \) set of image indices corresponding to artist \( k \), and \( G(\vec{x}_i) \) is the associated Gram matrix for each image. 
Then, to compute the distance between artists $k$ and $l$ on the basis of their Gram matrices, we simply take the $L_2$ norm of their element-wise difference. 
\begin{equation}
d_{\text{texture}}(\bar{G}_k, \bar{G}_l) = \left\| \bar{G}_k - \bar{G}_l \right\|_2^2.
\label{eqn:gram_distance}
\end{equation}

\section{Results}
We will verify this decomposition of style into a weighted average of symmetry similarity and textural similarity via a series of clustering experiments on a database of the 50 most well known artists throughout history. Our dataset consists of 1,700 paintings from 50 artists, evenly distributed by painter, and sampled randomly from a larger database of 17,000 paintings. We perform hierarchical clustering on these 50 artists on the basis of the style and texture features. In their 2012 paper, \citet{shamir_2012} perform a similar experiment detailing how hierarchical clustering based on a handpicked set of features can replicate groupings of artists based on shared style, period, and movement. In a similar spirit, we perform three experiments to empirically validate our definition. In particular, we demonstrate that when computing similarity with respect to both the Lie algebra generators and the Gram matrices we achieve a clustering of artists which aligns well with art-historical consensus. In the interest of transparency and reproducibility we state our assumptions about the ``ground truth'' of each artist's stylistic affiliation in the following section. A more complete breakdown of the movements to which each artist belongs may be found in Appendix A.


\subsection{Art Historical Ground Truth}

The ``Best Artworks of All Time''~\cite{icaro_best_artworks_2019} dataset comes with pre-labeled ``styles.'' However, many artists are tagged with multiple styles and some styles are redundant. We assigned each artist to one and only one artistic movement and combined overlapping categories. \footnote{In the end our categories are: Abstract art, Baroque Art, Byzantine Art, Cubism, Expressionism, Impressionism, Northern Renaissance, Pop Art, Post Impressionism, Primitivism, Renaissance, Romanticism, Surrealism, and Symbolism.} We now provide a brief overview of the genealogy and qualities of each category, which will prove important later, when interpreting the results of hierarchical clustering. Chronologically, the oldest movement is Byzantine Art, which is primarily religious in theme, notable for its bright colors, and departure from the naturalism of the earlier Classical period. \cite{Cormack2018hist} Next, Renaissance painters returned to the elegance of Classical forms, including a renewed focus on the human body. \cite{johnson2005ren} The Baroque movement, born of the Renaissance, emphasized the play of light and shadow in service of highly emotional compositions full of movement and tension. \cite{scheyer1937Baroque} Romanticism developed partially in response to the advent of rationalism, emphasizing the sublime chaos of nature. \cite{galitz2004romanticism} Emphasizing the beauty and light of the natural world, Impressionism represented a turning point in the history of art: suddenly, how a scene felt became just as important as how it looked. \cite{Johnson1977impress} In the years that followed, Post-Impressionists transformed the natural scenes which had captured the Impressionists, emphasizing artificial color palettes, and highly ''formal" composition. \cite{oxfordart_postimpressionism} Symbolism and Primitivism completed the transition from intellectual to emotional art, emphasizing dreamlike compositions. \cite{kaplan2003symb} Cubism took this emphasis on perspective  further, attempting to represent objects from multiple perspectives at once. From there, modern art followed: expressionism, abstract art, surrealism, and pop art characterized painting after the turn of the century, with artists moving fluidly between movements.

\subsection{Visualizing Learned Symmetries}
\begin{figure}
    \centering
    \includegraphics[width=.9\columnwidth]{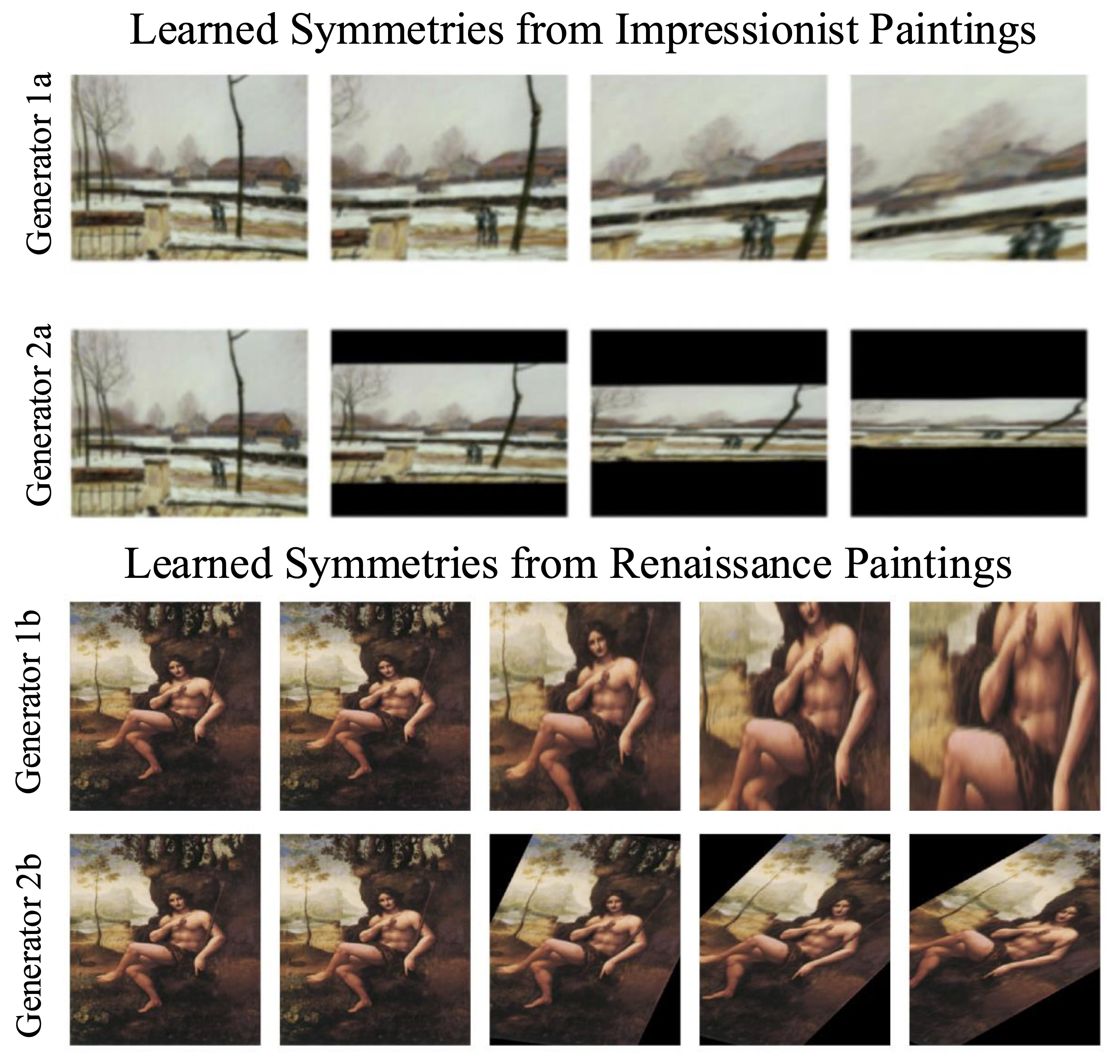}
    \caption{\textbf{Learned symmetry transformations from different artistic movements.} The symmetries were learned on Impressionist (top) and Renaissance (bottom) paintings.}
    \label{fig:generators}
\end{figure}

First, we demonstrate that our learned Lie generators produce interpretable transformations, when mapped back to the manifold. In Figure \ref{fig:generators}, we plot the action of four separate generators, estimated from two separate MLPs trained on datasets of Impressionist (top) and Renaissance paintings (bottom) respectively. We plot the action of these generators on a single randomly selected painting from each movement, with increasing scaling factors $t$ from left to right. Wansformations appear to have a notable simi style-specific. Impressionist paintings tend to depict horizontal, landscape scenes, often with well defined horizon lines, which corresponds to the horizontal compression when the action of generator 2a is applied. Conversely, Renaissance works tend to emphasize strong vertical lines, often because such paintings are vertical portraits, which corresponds to the zooming action of generator 1b, which emphasizes the central, vertical axis of the painting, which suggests that the Lie generators encode useful, interpretable information.

\subsection{Hierarchical Clustering}
\begin{figure}
    \centering
    \begin{subfigure}[h]{0.45\textwidth}
        \includegraphics[width=\linewidth]{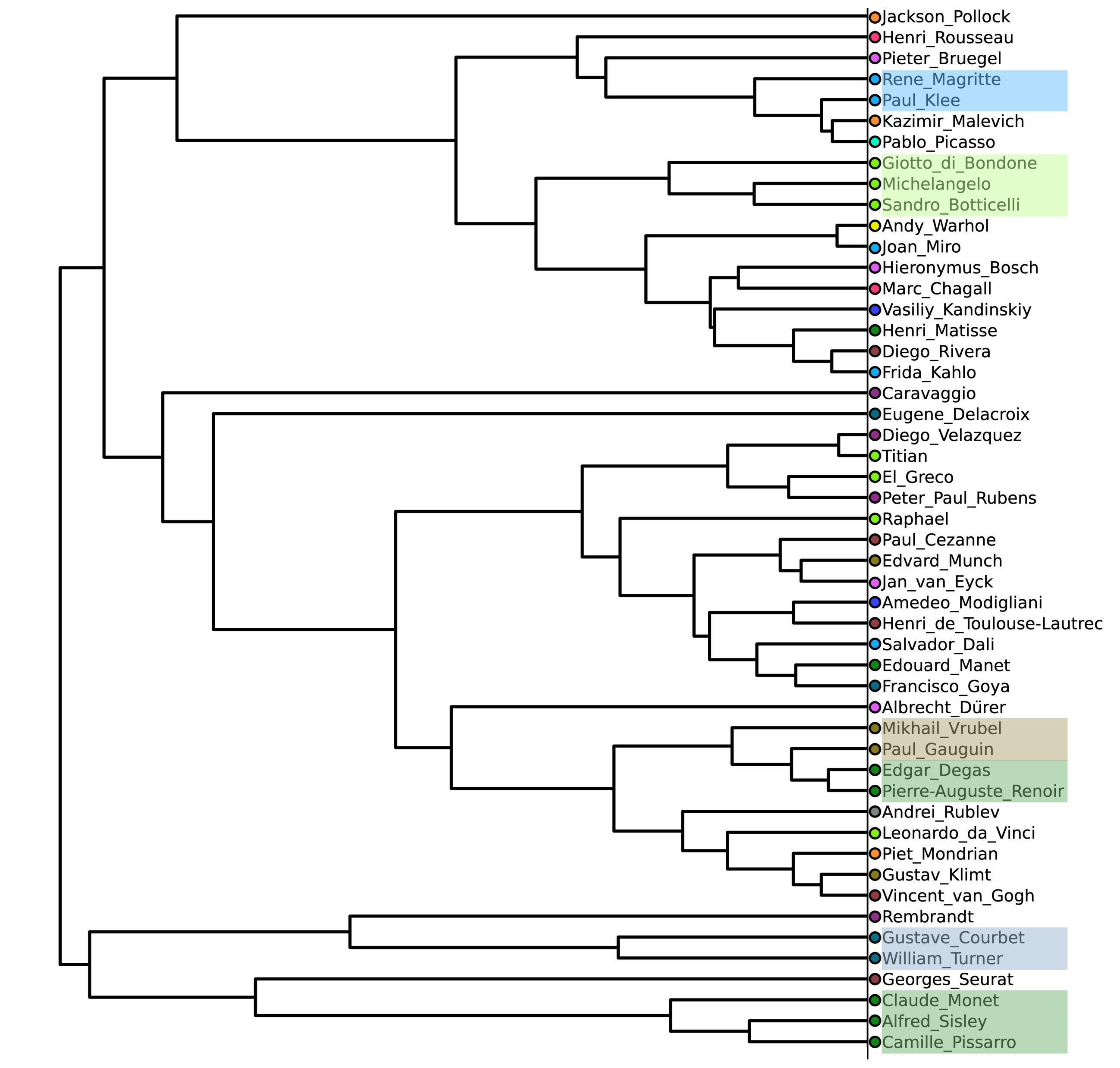}
        \caption{}
    \end{subfigure}

    \begin{subfigure}[h]{0.45\textwidth}
        \includegraphics[width=\linewidth]{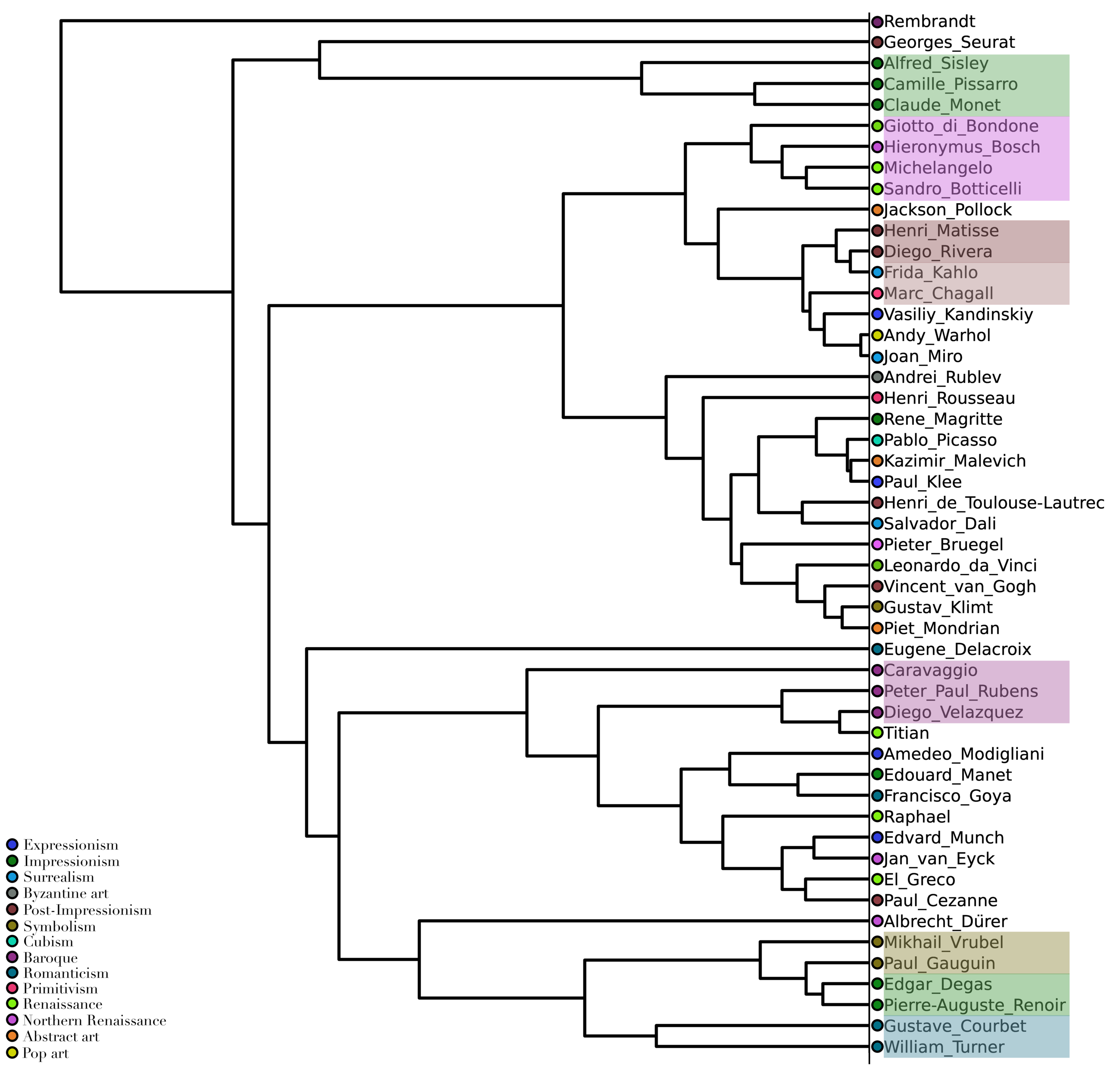}
        \caption{}
    \end{subfigure}
    \caption{\textbf{Phylogenic trees of artistic movements.} Hierarchical clustering of artists using (a) only Gram matrices and (b) both Gram matrices and Lie generators. We see that incorporating symmetry information in (b) yields a higher proportion of neighboring artists from the same art-historical movement (highlighted).}
    \label{fig:dendros}
\end{figure}

Following the example of \cite{shamir_2012}, we create hierarchical clustering dendrograms to visualize the similarity between artists. Specifically, we generate dendrograms using only the Gram matrices (Figure~\ref{fig:dendros}a) and using both the Gram matrices and Lie group generators combined (Figure~\ref{fig:dendros}b).

To construct Figure~\ref{fig:dendros}a, we begin by computing the average Gram matrix for each artist, as detailed in Section~\ref{sec:gram_distance}. 
Next, we calculate the distance between each pair of artists by taking the squared Euclidean norm of the difference between their average Gram matrices as detailed in Equation \ref{eqn:gram_distance}. 
This results in a \( 50 \times 50 \) distance matrix \( D \), where each element \( D_{kl} \) quantifies the stylistic difference between artist \( k \) and artist \( l \).
For Figure~\ref{fig:dendros}b, we improve our clustering by incorporating symmetry features through Lie group generators alongside the Gram matrices. We implement the distance measure between styles given by Equation ~\ref{eqn:combined} as follows. We define the distance between artists $k$ and $l$ to be a weighted average of their Gram distances and their Grassman distances, given in general by
\begin{equation*}
       \begin{split}
        D_{kl}^{\text{combined}} = & \, (1 - \lambda) \left\| \bar{G}_k - \bar{G}_l \right\|_2^2 \\
        & + \lambda d_{Gr}\left(\underset{i \in \{1, \ldots, 4\}}{\text{span}}(\mathfrak{h}^k_{i}), \underset{j \in \{1, \ldots, 4\}}{\text{span}}(\mathfrak{h}^l_{j})\right) 
    \end{split}
\end{equation*}
where the first term comes from Equation~\ref{eqn:gram_distance} and the second is given by the Grassman distances between the symmetry subspace of artist $k$ and the symmetry subspace of artist $l$, given by $\text{span}(\mathfrak{h}^k_{1}, ..., \mathfrak{h}^k_{4})$, and $\text{span}(\mathfrak{h}^l_{1}, ..., \mathfrak{h}^l_{4})$ respectively. 

We perform hierarchical clustering on this combined distance matrix using average linkage, resulting in the dendrogram depicted in Figure~\ref{fig:dendros}b. We see that this clustering, which incorporates both symmetry and texture, has a higher percentage of artists whose nearest-neighbors are a part of the same movement, when compared with the Gram matrix-only dendrogram of Figure \ref{fig:dendros}a. In particular, in the combined dendrogram, $40 \%$ of artists are ``closest'' to at least one artist of the same movement, as described in Appendix A, while in the Gram matrix only computation, only $30\%$ of painters satisfy this constraint.
Adding global symmetry information seems to improve hierarchical clustering particularly for more contemporary painters like the Post-Impressionists, Primitivists, and Surrealists. In addition, even for artists whose nearest neighbors do not share a movement, the overall groupings appear to agree with both a chronological and intuitive sense of which artistic movements share features -- for instance Pollock, Matisse, Rivera, Frida Kahlo, Chagall, Kandinskiy, Warhol, and Miro are all modern artists known for their expressive use of bright color and focus on non-representational forms, and they only appear as neighbors when considering symmetry information in addition to texture.

\subsection{Bootstrapping for Phylogenic Trees}
We further verify the validity of the clustering by applying a well-known method in evolutionary biology to assess the ``confidence'' that we should assign to each clade, based `on the proportion of bootstrap trees showing that same clade.' \cite{efron1996bootstrap, suzuki2006pvclust}. We resample the paintings with replacement 1000 times. Then, for each of those samples, we recompute our distance matrix and re-cluster according to that resampled distance matrix. Finally, we count the proportion of samples that contained each ``cluster'' in the original dataset. We run this procedure on both our texture only distance matrix and on our symmetry and texture distance matrix. Intuitively, we can consider these clusters identified by the bootstrap method as those the model is ``most confident'' in. In the texture only bootstrapping, we observe six clusters which were identified as present in more than $95\%$ of bootstrapping trials. However, of these clusters, only four are reasonable, given our ground truth as described in Table 2 of the appendix. 
The other two clusters erroneously associate, in the first case, Andrei Rublev (a Byzantine Painter), Klimt (a Symbolist), and Van Gogh (an Impressionist), and in the second case, Modigliani (an Expressionist) and Goya (a Romantic). In the symmetry and texture bootstrapping, we observed five different clusters present in more than $95\%$ of trials, only one of which is incorrect: again,  Modigliani and Goya. We hypothesize that this mistake comes from an over reliance on textural information, since it is present in both clusters. This accords with visual intuition, since both Goya and Modigliani make use of a muted, heavily brushed background for their compositions \cite{scheyer1937Baroque, galitz2004romanticism}. The clusters which the Lie and Gram matrix combined distance metric gave most support to were extremely chronologically interpretable. Warhol and Pollock, two of the best known American artists of the 20th Century were grouped together. Many of the great 20th Century European modernists were clustered together with high confidence as well: Diego Rivera, Frida Kahlo, Joan Miro, and Vasily Kandinsky formed a second cluster. Further results may be found in Appendix B.

\subsection{Mantel Tests}

In order to quantitatively verify that our method captures the underlying structure of the artist arrangements accurately, we apply the so-called Mantel test, which quantifies the degree of similarity between two distance matrices. ~\cite{sokal1995biometry, mantel1967detection}

Let $D^{(1)}$ and $D^{(2)}$ be two $n \times n$ distance matrices whose correlation we want to compute. Since distance matrices are symmetric, we vectorize the upper triangle, and denote these vectors as $d^{(1)}$ and $d^{(2)}$ respectively. The test statistic of interest, $r_M$, is then calculated as the Pearson correlation coefficient between $d^{(1)}$ and $d^{(2)}$: 
\begin{equation}
\scriptstyle
    r_M = \frac{\sum_{k=1}^{m} \left( d_k^{(1)} - \bar{d}^{(1)} \right) \left( d_k^{(2)} - \bar{d}^{(2)} \right)}{\sqrt{\sum_{k=1}^{m} \left( d_k^{(1)} - \bar{d}^{(1)} \right)^2} \sqrt{\sum_{k=1}^{m} \left( d_k^{(2)} - \bar{d}^{(2)} \right)^2}}
\end{equation}
where $d_k^{(1)}$ and $d_k^{(2)}$ are the $k$-th elements of the vectors $d^{(1)}$ and $d^{(2)}$, respectively, and $\bar{d}^{(1)}$ and $\bar{d}^{(2)}$ are the means of these vectors. Then, to perform hypothesis testing, we randomly permute the entries of $D^{(2)}$ to generate $D^{(2)}_{\text{permuted}}$ 1000 times. For each permutation of $D^{(2)}$ we compute $r_{M \text{ permuted }}$, and use these permuted test statistics to approximate the underlying null distribution of $r_M$. 
We compute the $p$-value according to where the observed $r_{M}$ falls in this null distribution. 

Here we choose to compare our distance matrices (our $D^{(2)}$) to three approximations of the ``ground truth'' in our data (our $D^{(1)}$). In particular combining information from our dataset \cite{icaro_best_artworks_2019} and from art historians \cite{oxfordart_postimpressionism, Johnson1977impress, akerman1962style, galitz2004romanticism, johnson2005ren, scheyer1937Baroque}, we construct three ``ground truth similarity matrices,'' from which we compute distance matrices, `Basic Ground Truth,' `Ground Truth,' and `Detailed Ground Truth.' To compute Basic Ground Truth, we assign two artists a similarity of 1 if two artists belong to the same movement (as detailed in Table 2 of the appendix) and 0 otherwise. To compute Ground Truth, we employ a few well-substantiated assumptions based on clear genealogical relationships across artistic movements. We set the the similarity between artist $i$ and artist $j$, the $(i,j)$th entry in our similarity matrix, to be 1 if the artists are the same. If artist $i$ and artist $j$ belong to the same movement their similarity is $0.75$. If one painter is a Renaissance painter and the other is a Northern Renaissance painter, we assign the similarity to b $0.5$. Similarly, if one painter is a Impressionist and the other is a Post-Impressionist, we assign the similarity to b $0.5$. Finally, we construct our most detailed version of our ground truth distance matrix. We compute Ground Truth, then we assign a similarity of 0.25 to pairs where a) one artist is a Baroque painter and the other is a Renaissance painter b) one artist is an abstract painter and the other is an expressionist and c) one artist is a pop artist and the other is an abstract artist. We call this matrix `Detailed Ground Truth.' 

We summarize our findings in Table~\ref{tab:pvals}, where it seems that our Lie+Gram matrix clustering produces comparable or better p-values than the Gram matrices alone for every ground truth matrix. These results provide significant empirical validation for our proposed definition, which is both computationally tractable and highly descriptive of the stylistic associations of each movement. 

\begin{table}[h!]
	\centering
	\begin{tabular}{cccc}
	   \toprule
        \text{ } &\textbf{Basic GT} & \textbf{GT} & \textbf{Detailed GT} \\
        \midrule
        Gram Dist  & 0.007 & 0.00332 & 0.008\\
        Lie+Gram Dist & 0.002 & 0.003404 & 0.005 \\
	   \bottomrule
    \end{tabular}
    \caption{$p$-values of the Mantel test statistic.}
    \label{tab:pvals}
\end{table}

\section{Conclusion}
Our work proposes and empirically validates a new definition of artistic style in terms of the symmetries and texture associated with an image. By treating Lie algebra generators as a proxy for the symmetries of an image and Gram matrices as a description of its texture, we identify a quantitative method for determining the style of an image.  By validating our approach with clustering experiments, we demonstrate that this dual focus on symmetry and texture yields groupings that align with art-historical consensus. Through clustering experiments, we demonstrate that this combined approach not only aligns closely with art-historical assignments of artists to specific movements, but also enhances the interpretability of stylistic distinctions across movements and individual artists. This framework sets the stage for more theoretically grounded and  perceptually aligned methods in style transfer, providing a promising avenue for both art analysis and generative modeling. We hope that a more precise notion of style will help advance both the interpretability and the perceptual accuracy of new style-transfer methods in future work.  \\

\clearpage
\setcounter{page}{1}
\maketitlesupplementary
\onecolumn
\section{Appendix}
\subsection{Art Historical Context}

In Table 2, we assign each of the fifty artists to a unique art movement throughout history. These categories formed the basis for our similarity analysis detailed in Section 5. Figure Four explains the chronological arrangement of art movements from Byzantine Art to Impressionism. Until the late 1800s, artistic movements were mostly determined chronologically and geographically. After the Impressionists, style became more fluid and the pace of change rapidly increased. In Figure 5, we depict the relationship between artistic styles after the 1800s. Many styles developed rapidly in parallel after Impressionism and artists often moved fluidly between movements. These images and artists were chosen because they are roughly indicative of the style of the movements.  
\label{sec:AppendixA}
\begin{table*}[htbp]
\centering
\caption{Artists by Artistic Movements}
\label{fig:artist_movement_table}
\begin{tabularx}{\textwidth}{>{\raggedright\arraybackslash}p{4cm} X}
\toprule
\textbf{Artistic Movement} & \textbf{Artists} \\
\midrule
Abstract Art &
Piet Mondrian, Kazimir Malevich, Jackson Pollock \\
\addlinespace
Baroque &
Peter Paul Rubens, Caravaggio, Diego Velazquez, Rembrandt \\
\addlinespace
Byzantine Art &
Andrei Rublev \\
\addlinespace
Cubism &
Pablo Picasso \\
\addlinespace
Expressionism &
Amedeo Modigliani, Vasiliy Kandinsky, Edvard Munch, Paul Klee \\
\addlinespace
Impressionism &
Claude Monet, Edouard Manet, Pierre-Auguste Renoir, Alfred Sisley, Edgar Degas, Camille Pissarro \\
\addlinespace
Northern Renaissance &
Hieronymus Bosch, Albrecht Dürer, Pieter Bruegel, Jan van Eyck \\
\addlinespace
Pop Art &
Andy Warhol \\
\addlinespace
Post-Impressionism &
Vincent van Gogh, Henri Matisse, Henri de Toulouse-Lautrec, Paul Cézanne, Georges Seurat, Diego Rivera \\
\addlinespace
Primitivism &
Marc Chagall, Henri Rousseau, Paul Gauguin\\
\addlinespace
Renaissance &
Giotto di Bondone, Sandro Botticelli, Leonardo da Vinci, Raphael, Michelangelo, Titian, El Greco \\
\addlinespace
Romanticism &
Francisco Goya, William Turner, Eugène Delacroix, Gustave Courbet \\
\addlinespace
Surrealism &
René Magritte, Salvador Dalí, Frida Kahlo, Joan Miró \\
\addlinespace
Symbolism &
Gustav Klimt, Mikhail Vrubel, Edvard Munch \\
\bottomrule
\end{tabularx}
\end{table*}

\begin{figure}
    \centering
    \includegraphics[width=.9\linewidth]{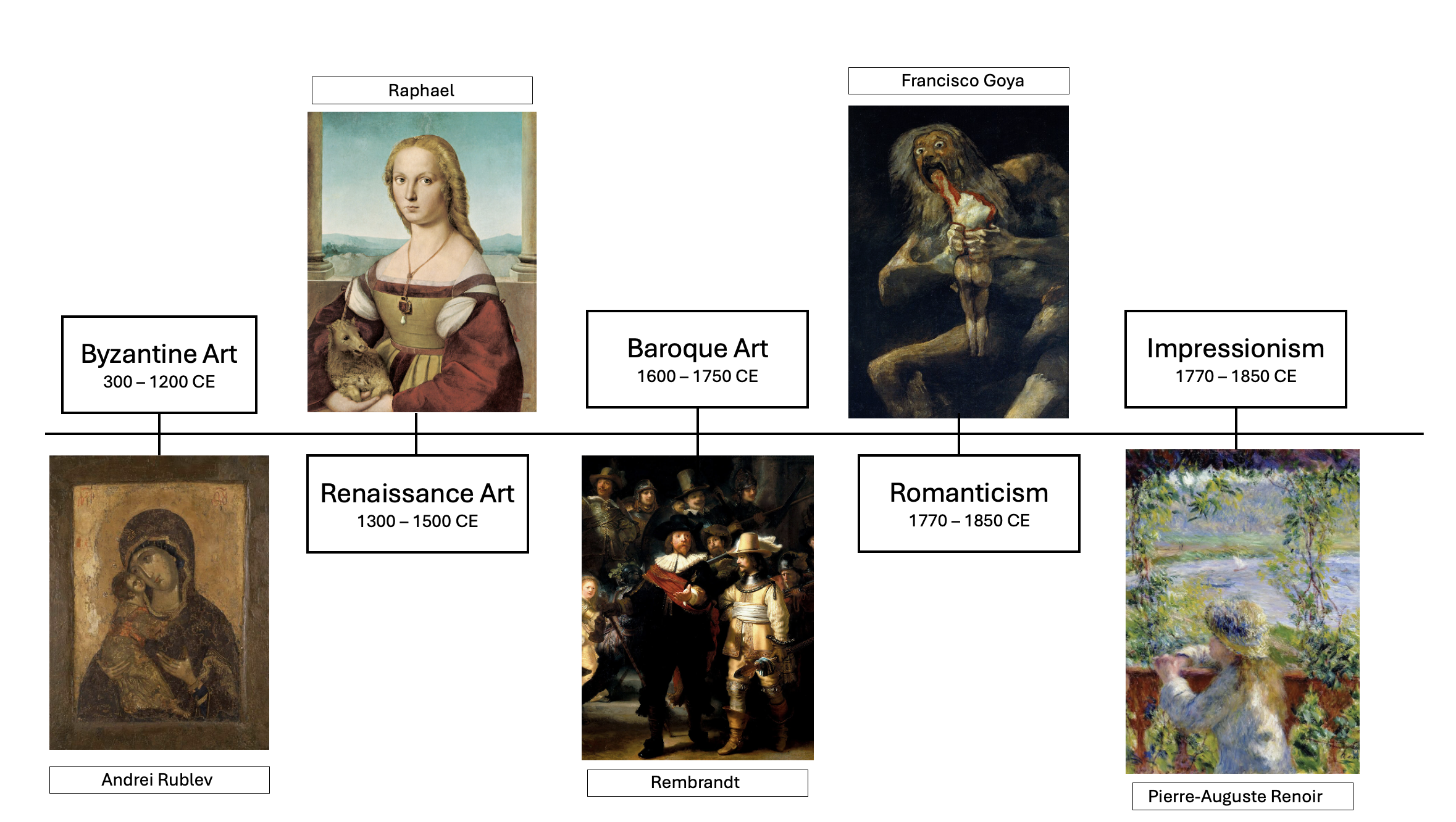}
    \caption{Timeline of Art Movements:}
    \label{fig:enter-label}
\end{figure}

\begin{figure}
    \centering
    \includegraphics[width=0.7\linewidth]{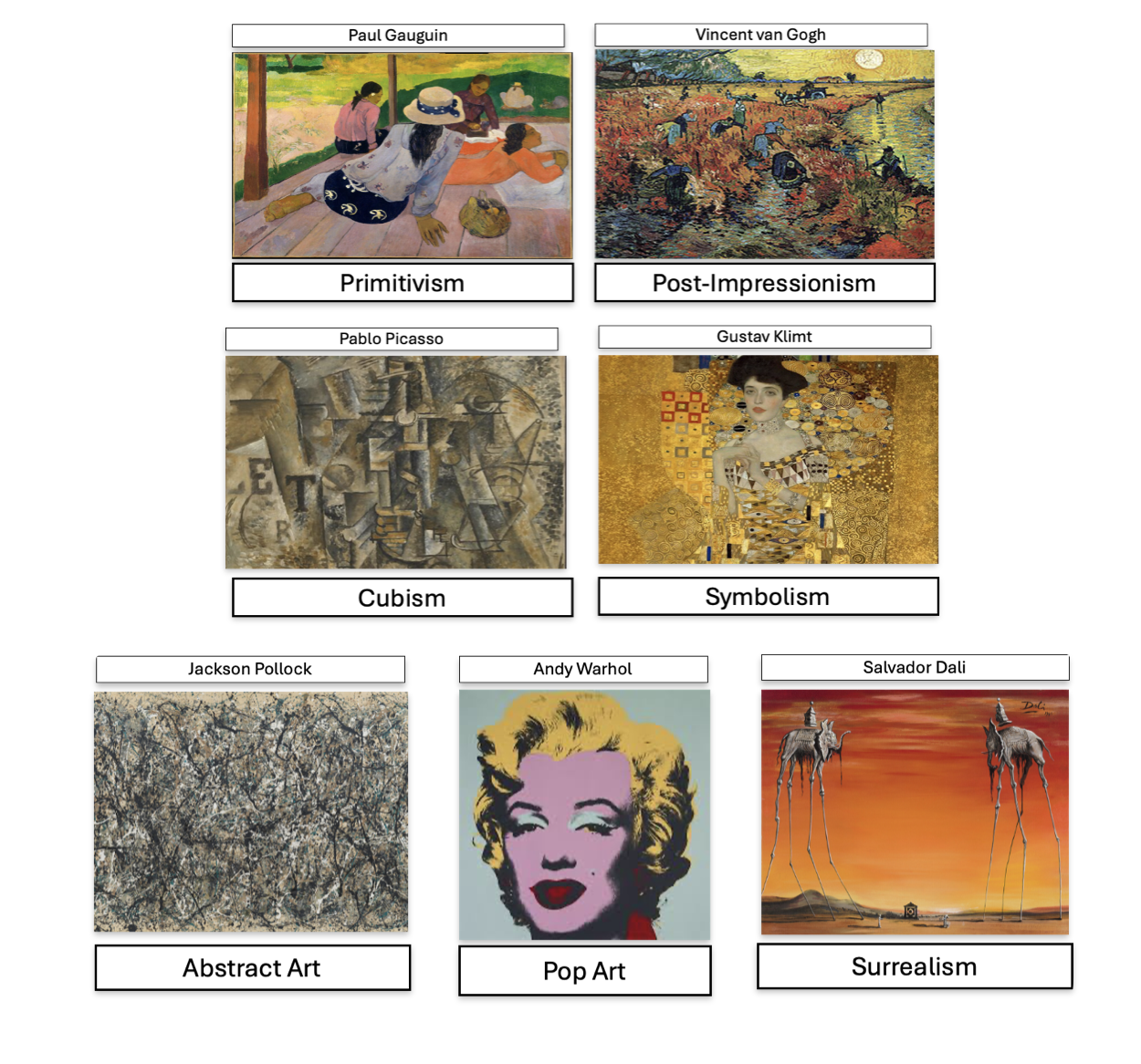}
    \caption{Artistic Movements after Impressionism, in roughly chronological order from top to bottom.}
    \label{fig:enter-label}
\end{figure}

\begin{figure}
    \centering
    \includegraphics[width=1\linewidth]{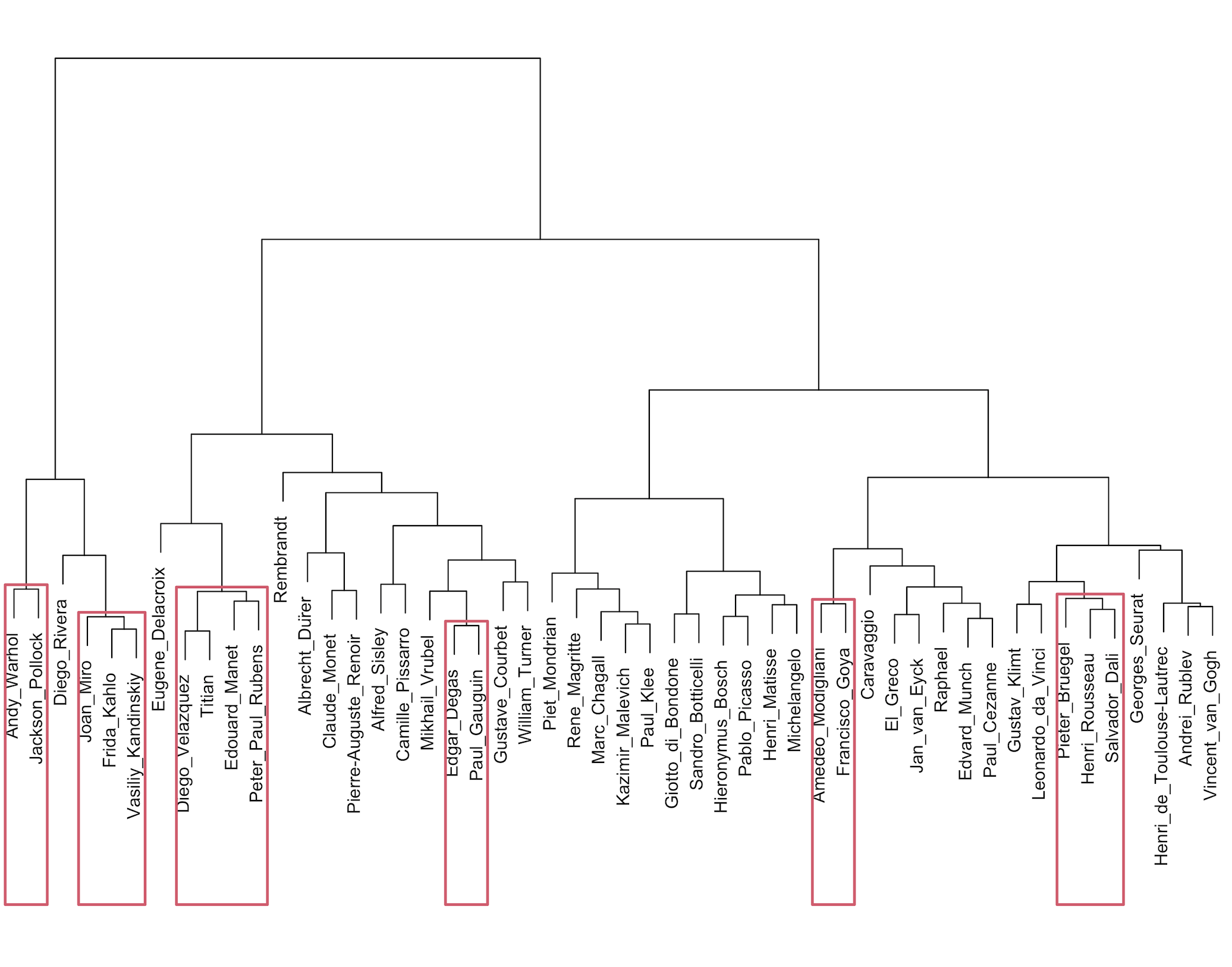}
    \caption{\textbf{PV Clustering Based on Lie Generators and Gram Matrices:} This is a diagram of the clustering produced by the PV Clustering Algorithm run on the similarity matrix generated by an even weighting of the Lie similarities and Gram matrix similarities. The red boxes represent "clusters" that appear in more than 90\%  of bootstrap trials.}
    \label{fig:enter-label}
\end{figure}
\subsection{Bootstrapping for Phylogenic Trees}
\twocolumn
%

\pagebreak
{
    \small    \bibliographystyle{ieeenat_fullname}
    \bibliography{main}
}


\end{document}